\documentclass{Interspeech2024}

\interspeechcameraready

\usepackage[table,dvipsnames]{xcolor}
\usepackage{todonotes}
\usepackage{amsmath,amssymb,graphicx}
\usepackage{tabularx}
\usepackage{multirow}
\usepackage{ctable,booktabs}
\usepackage[inline]{enumitem}

\usepackage[linesnumbered, ruled, ruled, noend]{algorithm2e}
\usepackage{setspace}
\usepackage{etoolbox}
\usepackage{xspace}

\newcommand{\sysname}{SALSA\xspace}
\SetKwInOut{Input}{Input}  
\newlength\mylen

\SetKwInOut{Output}{Output}

\SetCommentSty{mycommfont}

\usepackage{defs}
\newcommand{\vocab}{\mathcal{V}}

\newcommand{\vC}{{\vek{C}}}

\newcommand{\cL}{\mathcal{L}}
\newcommand{\vH}{{\vek{H}}}

\def\vx{{\bm{x}}}
\def\vy{{\bm{y}}}
\newcommand{\DA}{d}
\newcommand{\DL}{d^L}
\newcommand{\vocabL}{\vocab^L}
\newcommand{\state}{\vs}
\newcommand{\stateL}{\state^L}

\newcommand{\iI}{i^{\text{th}}}
\newcommand{\iJ}{j^{\text{th}}}

\newcommand{\iT}{t^{\text{th}}}
\newcommand{\iR}{r^{\text{th}}}

\title{SALSA: Speedy ASR-LLM Synchronous Aggregation}
\name[affiliation={*1,2}]{Ashish}{Mittal}
\name[affiliation={*2}]{Darshan}{Prabhu}
\name[affiliation={2}]{Sunita}{Sarawagi}
\name[affiliation={2}]{Preethi}{Jyothi}

\address{$^1$IBM Research,  $^2$IIT Bombay}
\email{arakeshk@in.ibm.com, \{darshanp, sunita, pjyothi\}@cse.iitb.ac.in}


\begin{document}

\maketitle

\begin{abstract}
Harnessing pre-trained LLMs to improve ASR systems, particularly for low-resource languages, is now an emerging area of research. Existing methods range from using LLMs for ASR error correction to tightly coupled systems that replace the ASR decoder with the LLM. These approaches either increase decoding time or require expensive training of the cross-attention layers. We propose \sysname, which couples the decoder layers of the ASR to the LLM decoder, while synchronously advancing both decoders. Such coupling is performed with a simple projection of the last decoder state, and is thus significantly more training efficient than earlier approaches. A challenge of our proposed coupling is handling the mismatch between the tokenizers of the LLM and ASR systems. We handle this mismatch using cascading tokenization with respect to the LLM and ASR vocabularies. We evaluate SALSA on 8 low-resource languages in the FLEURS benchmark, yielding substantial WER reductions of up to $38\%$.
\def\thefootnote{*}\footnotetext{These authors contributed equally to this work.}\def\thefootnote{\arabic{footnote}}

\end{abstract}

\section{Introduction}

Automatic speech recognition (ASR) systems, whether they are cascaded or end-to-end, have been shown to benefit from both a strong acoustic model and a strong language model. When the acoustic model is inadequate, particularly for low-resource languages with limited access to labeled speech, the language model can offer effective supplementary support~\cite{xu2013cross,gandhe2014neural}. With the advent of pretrained large language models (LLMs) and their superior text modeling abilities, there is growing interest in leveraging LLMs to improve ASR performance~\cite{wu2023decoder,touvron2023llama,fathullah2023prompting}. This LLM-ASR integration could be particularly beneficial for low-resource languages on which the ASR model underperforms and for which the LLM model has sufficient base capabilities.

There are broadly three ways in which LLMs have been leveraged for ASR in recent work.
\begin{enumerate}
\item ASR error correction: The ASR system generates $N$-best lists and the LLM  rescores them. This could be achieved by prompting the LLM with the $N$-best list and attending to the speech encoder~\cite{radhakrishnan2023whispering,chen2024hyporadise}.
\item Speech in-context learning: Pretrained LLMs are additionally instruction-tuned with speech inputs enabling a tight coupling between a speech encoder and the LLM to support multiple speech tasks including ASR~\cite{wu2023decoder,zhang2023speechgpt,pan2023cosmic}.
\item Deep LLM-fusion: The LLM replaces the decoder of an encoder-decoder ASR system via gated cross-attention~\cite{fathullah2023prompting}.
\end{enumerate}
All these prior approaches are computationally expensive with high training overhead in the deep LLM-fusion and speech in-context learning paradigms due to fine-tuning, and high decoding latency due to second-pass rescoring in ASR error correction. In this work, we propose a lightweight alternative called \sysname%
\footnote{Code for \sysname is available at \url{https://github.com/csalt-research/salsa}.
}
that offers a deeper integration of LLMs with ASR beyond shallow fusion while incurring a low training overhead. To the best of our knowledge, we are also the first to show the utility of LLMs for ASR of a diverse set of low-resource languages; all prior work in this area has focused on English ASR.

In \sysname, we keep the pretrained ASR and LLM backbone architectures frozen and only train feedforward projection layers that couple the ASR decoder layers to the LLM decoder layers.  \sysname\ requires that both ASR and LLM decoders move forward in tandem albeit having different tokenizations.  The LLM autoregressively predicts the next token, and as soon as a valid ASR tokenizable text is formed, advances the ASR decoder with the predicted text. With each synchronized step, the learned projection layers act on the last state of the ASR decoder and are added as a residual connection to the LLM decoder layer's representations. 
%We address this mismatch in tokenization by successively detokenizing the LLM prediction using its vocabulary and retokenizing using the ASR vocabulary to advance the ASR decoder.  
%
The advantage of such a coupling is that the ASR's cross-attention layers are retained, and a simple projection from the ASR decoder states to the LLM decoder states suffices.  We will show that such a coupling leads to significantly faster training of the coupling parameters than existing approaches such as Whispering-Llama~\cite{radhakrishnan2023whispering}. 

\sysname can be used to integrate any pretrained decoder-only LLM with a pretrained encoder-decoder ASR model using small amounts of labeled speech in the target languages. By projecting just the last state of the ASR decoder at each LLM decoding step, we bypass the need for learning cross-attention modules, thus making \sysname significantly more parameter-efficient than existing approaches. We implement \sysname using a pretrained Whisper ASR model~\cite{whisper} and LLama-2~\cite{touvron2023llama}. On eight diverse languages in the FLEURS \cite{conneau2023fleurs} benchmark, we obtain a significant $16\%$ on average and up to a maximum of $38\%$ relative reduction in WER compared to parameter-efficient finetuning.  In contrast, existing fusion approaches that rely on $n$-best lists from ASR perform much worse for low resource languages.
\looseness=-1

%The text encoder $\cM_L$ converts each text prefix $\vy_u=y_1,\ldots,y_{u-1}$ to the  %$\vg_1,\ldots,\vg_U$. %The most common choice for a text encoder is an RNN. 
% followed by a softmax layer that outputs the distribution $\pmodel(y|\vg_u, \vh_t)$ over vocabulary $\vocab$ plus the blank symbol $\varnothing$.  
\section{Related Work}
\xhdr{LM adaptation in ASR} Traditional ASR systems consist of decoupled acoustic and language models~\cite{mohri2002weighted} which enables easier adaptation to a target domain~\cite{hori2003language,bellegarda2004statistical,neubig2009wfst,gangireddy2016unsupervised,park2010improved,xu2018pruned}. For end-to-end ASR models, the most popular approach to integrate an external LM is shallow fusion and its variants, where an external LM is log-linearly interpolated with the ASR model~\cite{kannan2018, mcdermott2019density,meng2021internal,meng21_interspeech,udagawa2022effect, mittal2023situ, mittal2023speech}. 

\xhdr{LLM adaptation in ASR} With the rapid progress on various natural language tasks using LLMs, their integration with ASR models is emerging as an area of significant interest. One of the early application of LLMs was to use them for ASR error correction by providing an $n$-best list and a prompt to generate the correct prediction \cite{chen2024hyporadise,dighe2023leveraging, yang2023generative, ma2023can}. These solutions are heavily reliant on the ASR outputs and will not fare well on low-resource languages owing to large errors in the $n$-best predictions.
%that work on the output from the ASR model are not faithful to the underlying audio.

\xhdr{Deep LLM integration with an ASR system}
An active area of research is to integrate the audio modality within an LLM. Whispering-Llama ~\cite{radhakrishnan2023whispering} learns adapter layers~\cite{zhang2023llama, gao2023llama} to cross-attend to the audio features and prompts with an $n$-best list to improve English ASR. Another popular approach is to provide the output of an audio encoder directly as an input to decoder-only LLMs. Full fine-tuning is done to semantically map the acoustic features with the underlying textual features within an LLM \cite{wu2023decoder,fathullah2023prompting, pan2023cosmic, lai2023instruction, shu2023llasm, ma2024embarrassingly}. The closest to our work is \cite{chen2024s}, that does a late integration of the ASR and LLM decoders. However, they work with $n$-best lists (that is not conducive to low-resource languages) and finetune the ASR decoder using the LLM's tokenizer to match the LLM vocabulary. \sysname does not need such an additional finetuning step and can work with different ASR/LLM tokenizations.  

\section{Our Approach: \sysname}
\begin{figure}[t!]
    \centering
    \includegraphics[width=0.99\linewidth]{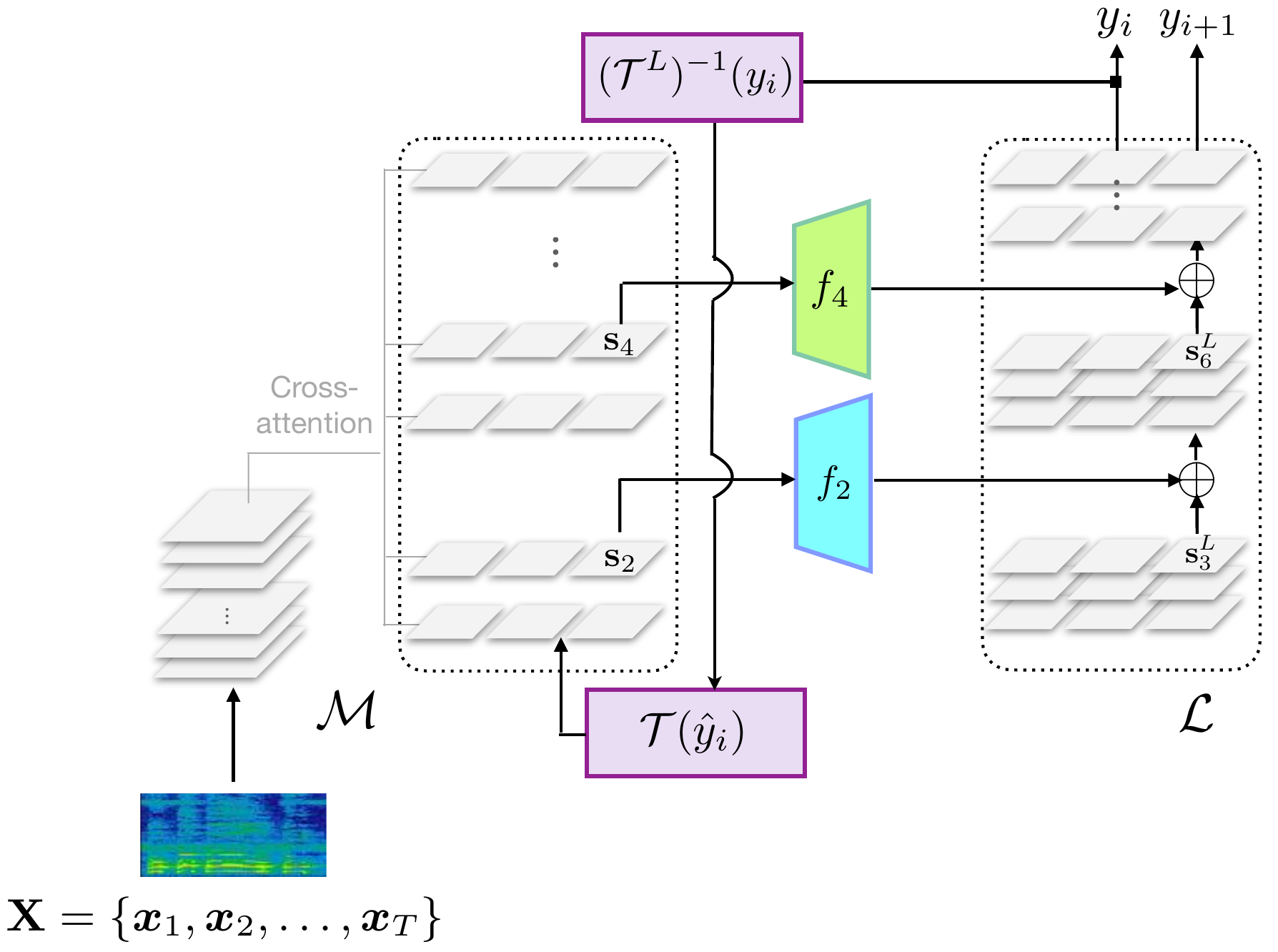}
    \caption{\sysname: Overall schematic illustrating the coupling of the ASR model $\cM$ and the LLM $\cL$ using select projection layers. For the sake of simplicity, we will assume $y_i$ corresponds to a single token in this illustration.}
    \label{fig:salsa}
\end{figure}
The input to our model is a trained ASR model $\cM$, an LLM model $\cL$, and a  small set of labeled audio-transcript pairs $D=\{(\vx^n,\vy^n):n=1\ldots N\}$ in a target low-resource language.  We assume that the LLM has been pre-trained with significantly more text data in the target language, compared to the speech transcription data used in the ASR model.  % We present a method of verifying this assumption before proposing to use an LLM to improve an ASR model in Section~\ref{sec:modelselection}.  
We first present a brief background of the ASR and LLM models.
\looseness=-1

\xhdr{Background: ASR model}
We assume the ASR system $\cM$ is an encoder-decoder model as used in state-of-the-art ASR systems like Whisper~\cite{whisper}. 
The encoder $\cM_S$ converts an input audio $X$ comprising of $T$ frames  $\vx_1,\ldots,\vx_{T}$ into their latent vectors $\vh=\vh_1,\ldots,\vh_T$.
%using an RNN~\cite{cho2014properties} or Transformers \cite{vaswani2017attention} or Conformer~\cite{gulati2020conformer}. 
%
The transcript $\vy$ is generated auto-regressively as per the vocabulary $\vocab$ and a tokenization algorithm $\cT$.  Let  $\vy_u$=$y_1,\ldots,y_u$ denote the token sequence of the transcript generated so far.  
The decoder takes as input the prefix $\vy_u$ and cross-attends on the audio states  $\vh_1,\ldots,\vh_T$  to generate the next token via multiple self-attention layers.  Let $\DA$ denote the number of decoder layers.  Let $\state_{\ell} \in \Re^m$ denote the vector output from the decoder at layer $\ell \in [1,\DA]$. 

\xhdr{Background: LLM model}
We assume the  $\cL$ is a decoder-only model consisting of $\DL$ layers.
Conditioned on an instruction prompt and the partially generated text, the LLM also generates the next token autoregressively. 
Let $\vocabL$ denote the vocabulary of the LLM and let $\cT^L$ denote its tokenizer.  In general, both $\vocabL$ and $\cT^L$ could be different from $\vocab$ and  $\cT$ respectively of the ASR model. At any step $t$ of generation, let $\stateL_{\ell} \in \Re^{m^L}$ denote the decoder output from each layer $\ell \in [1,\DL]$.   Finally, a  softmax yields $P(y|\stateL_{\DL})$ where $y \in \vocabL$.

\vspace{0.1in}
%\subsection{\sysname}
Our method of coupling the ASR model $\cM$ with the  LLM model $\cL$ is to just add a lightweight bridge from the latest state of the ASR decoder to the LLM decoder state as shown in Figure~\ref{fig:salsa}. We choose a subset $F \subseteq \{1\ldots,\DL\}$ of LLM decoder layers to connect to $|F|$ different ASR decoder layers.  Let $I(\ell)$ denote the index of the ASR decoder layer to which a LLM decoder layer $\ell \in F$ is connected.  We use $f_{\ell}: \mathbb{R}^{m} \rightarrow \mathbb{R}^{m^L}$ to denote the feed forward network for the $\ell$th layer. The output of this function is used to add a residual connection to the LLM's decoder output at layer $\ell$ as: 
\begin{equation}
\label{eq:couple}
    \stateL_\ell = \stateL_\ell + f_\ell(\state_{I(\ell)})
\end{equation}
The set of parameters over all $f_\ell:\ell \in F$ is denoted as  $\theta_C$.  We experimented with different methods of choosing the set $F$ and $I$.  Our default method, if size of $F$ is $k$, is to make $F$ every $d^L/F^{~\text{th}}$ layer of $\cL$, and $I(F)$ be the corresponding $d/F^{~\text{th}}$ decoder layer of $\cM$. We will present ablation on size of $F$, and alternative methods of choosing $F$.
%\footnote{For simplicity, we employ a symmetric configuration where the projection layers are placed at equal intervals. A more complex setup, with either shared projection layers or assymetrically spaced projection layers can also be explored.}. We will use $I(\ell)$ to denote the ASR decoder layer to which the $\ell$th layer of the LLM model is coupled. 
% Given that $\cL$ is mainly responsible for generating the transcriptions, 
%The role of the projection layer $f_{\ell}$ is to incorporate the output from  decoder layer $I(\ell)$ of $\cM$ into the output of  decoder layer $\ell$ of $\cL$. 
%

\xhdr{Transcript generation}
The LLM model $\cL$  generates the next token auto-regressively.  The LLM's softmax layer yields a distribution $P(y|\stateL_{\DA})$ over its vocabulary $\vocabL$.  From this, the sampled $y_{t+1}$ forms the next generated token.   In general, this token may not be  valid text recognized by the ASR tokenizer $\cT$. Often for low resource languages the tokenizers for $\cL$ and $\cM$ can use different multi-token sequences to encode a character in the target language.  So, the LLM keeps generating tokens until a valid text piece recognizable by ASR's $\cT$ is formed.  As soon as  the LLM decodes a complete character sequence (which can be a single character or a set of characters), the just generated text is re-tokenized with the ASR's tokenizer  to convert into a sequence of tokens that can be understood by $\cM$. Finally, the decoder state of $\cM$ is advanced with the newly predicted sequence of tokens. This updated decoder state is then used by $\cL$ in its subsequent decoding iterations. The synchronous invocation of different decoders with different vocabularies is a distinctive aspect of our approach.  Algorithm~\ref{alg:inference} gives an overview of decoding in \sysname.
\looseness=-1

\begin{algorithm}[h]
\caption{\sysname\ Decoding Algorithm}
\label{alg:inference}
\DontPrintSemicolon
\Input{ 
   ASR Model  $\cM$,  LLM Model $\cL$, Audio: $X$\\
    Tokenizers $\cT$ of ASR, $\cT^{L}$ of LLM \\
     $f_\ell, F, I:$ coupling  model specification.
%    $\text{max}_{l}$: Max number of tokens to be predicted \\
%    $\texttt{isValid}(.)$: Checks if there exists non \\ 
%    \hspace{1.2cm} utf-8 characters and not empty \\
}
\Output{ 
    Transcript $\{ \vy_0,\ldots,\vy_u \}$
}
\setstretch{1.2}
    $\vh$: ASR Encoder State ($X$) \\
    $\state$ = ASR Decoder State initialized with SOT token \\
    $\stateL$ = Initial LLM Decoder State \\
pi = 0, $\vy$ = [] \\
\For{$i \in \{ 1 \ldots $ max estimated tokens in $X$\}}{
    $\stateL_\ell = \stateL_\ell + f_\ell(\state_{I(\ell)}),~~~\forall \ell \in F$\\
    $y_i$, $\stateL$ = $\cL$($\stateL$) \\
    $\vy$.append($y_i$) \\
    textY = $(\cT^{L})^{-1}(\vy)$ \\
%    \If{ \text{\normalfont \texttt{isValid}~(}\text{\normalfont text)} }{
\If{textY ends with a valid utf-8 character}{
        new\_tokens = $\cT$(textY[pi:]) \\
        $\state$ = $\cM$($\vh$, $\state$, new\_tokens) \\
        pi = \texttt{len}(textY)
    }
    If {$y_i$ == $\cT^{L}$.eos } break;
}
\Return{ $\vy$ }
\end{algorithm}
%
%

%This check ensures that $\cL$ predicts all the tokens that can be encoded to a character sequence before updating the state of to $\cM$ model. 

%The decoders of the ASR and LLM progress almost synchronously and are conditioned on the same generated text so far. However, because of the potential difference in the tokenization of the two models, the exact token sequence may be different. 

%We elaborate on how we handle this difference in the training and inference steps elaborated in Sections~\ref{subsec:methodology:training} and ~\ref{subsec:methodology:inference}  respectively.

%Our proposed approach aims at learning a composite model $\cC$ that combines the representation power and language modelling capability of $\cM$ and $\cL$. To combine both these models,

\xhdr{Training} \label{subsec:methodology:training}
We train only the coupling parameters $\theta_C$ using the limited training dataset $D$ in the target language. The parameters of both the ASR and LLM models are kept frozen. During training, an added detail is to tokenize a gold transcript $\vy$ first using the LLM's tokenizer $\cT^L$, then tokenize using the conditional tokenization of the ASR model as elaborated above, and finally remember for each index in the LLM's token sequence, the aligned sequence in the ASR tokenization. Otherwise, training proceeds using teacher forcing as in normal encoder-decoder models with cross entropy loss on the probabilities produced by LLM's decoder.

In spite of the need to handle such differences in tokenization, we found two advantages of such a coupling:  (1) Since the ASR decoder has already been trained for cross-attention on the encoded audio, the LLM decoder does not need to be retrained for this task.  Instead, since both the ASR and LLM decoders are auto-regressive, the LLM only needs to  consult the last ASR decoder state. (2) Since the LLM is assumed to be better at modeling the target language, the final text generation is with the LLM's decoder with a residual connection to the ASR decoder.

\section{Experimental Setup}
\begin{table*}[t!]
\centering
\caption{Comparison of WERs (\%) of \sysname using LLaMA-7B and LLaMA-13B models against Whispering-LlaMA and different variants of Whisper model on eight languages of \textsc{Fleurs}. Numbers in bold without \colorbox{green!20}{$\phantom{x}$} denote the best across baselines and with \colorbox{green!20}{$\phantom{x}$} denotes the best WER across all experiments. $\dagger$ indicates that the performance gains are statistically significant at $p < 0.001$.}
\label{table:res_asr}
\resizebox{\linewidth}{!}{
    \begin{tabular}{ l ccccccccc|c }
        \hline
        \hline
        
        % \multicolumn{1}{c|}{\multirow{2}{*}{\textbf{\small{Method}}}} & \multicolumn{2}{c|}{\textbf{\footnotesize{Overall}}} & \multicolumn{2}{c|}{\textbf{\footnotesize{Hindi}}} & \multicolumn{2}{c|}{\textbf{\footnotesize{Gujarati}}} & \multicolumn{2}{c|}{\textbf{\footnotesize{Marathi}}} & \multicolumn{2}{c|}{\textbf{\footnotesize{Telugu}}} & \multicolumn{2}{c|}{\textbf{\footnotesize{Tamil}}} & \multicolumn{2}{c|}{\textbf{\footnotesize{Malayalam}}} & \multicolumn{2}{c}{\textbf{\footnotesize{Punjabi}}} \\

        \multicolumn{1}{c}{\textbf{\scriptsize{Method}}} & \textbf{\scriptsize{\# params}} & 
        \textbf{\scriptsize{Gujarati}} &\textbf{\scriptsize{Hindi}} &  
        \textbf{\scriptsize{Malayalam}} & \textbf{\scriptsize{Marathi}} & 
        \textbf{\scriptsize{Persian}} &
        \textbf{\scriptsize{Punjabi}} &
         \textbf{\scriptsize{Tamil}} &  
         \textbf{\scriptsize{Telugu}} &
         \textbf{\scriptsize{Average}} \\

        \hline

        % \footnotesize{Whisper~(Large-v2)~\cite{whisper}} & -- & \footnotesize{64.4} &  \footnotesize{21.5} &  \footnotesize{102.7} &  \footnotesize{38.3} & \footnotesize{99.0} & \footnotesize{17.5} & \footnotesize{100.7} & \footnotesize{102.4} & \footnotesize{32.9} \\

        % \hline

        \multicolumn{10}{l}{\footnotesize{\textit{Our Baselines}~~\textcolor{gray}{(Reproduced using official repositories)}}}\\

        \hspace{1.5mm} \footnotesize{Whisper~(Large-v2)~\cite{whisper}} & -- 
        & \footnotesize{108.2} 
        &  \footnotesize{35.9} 
        &  \footnotesize{107.8} 
        &  \footnotesize{84.7} 
        & \footnotesize{35.8} 
        & \footnotesize{101.8} 
        & \footnotesize{48.0} 
        & \footnotesize{104.4} 
        & \footnotesize{78.3} \\ 
        % \hspace{4mm} \footnotesize{ w/ Greedy} & \footnotesize{22.7} &  \footnotesize{17.3} &  \footnotesize{28.0} &  \footnotesize{18.1} & \footnotesize{17.8} & \footnotesize{19.7} & \footnotesize{18.5} & \footnotesize{16.3} & \footnotesize{25.9} \\

        \hspace{4mm} \footnotesize{ w/ \textsc{LoRA} fine-tuning~\cite{lora}} & \footnotesize{15M} & \footnotesize{\textbf{55.7}} &  \footnotesize{\textbf{19.3}} &  \footnotesize{\textbf{54.9}} &  \footnotesize{\textbf{35.8}} & \cellcolor{green!20} \footnotesize{\textbf{16.9}~$\dagger$} &  \footnotesize{\textbf{46.7}} & \cellcolor{green!20} \footnotesize{\textbf{38.0}~$\dagger$} & \footnotesize{\textbf{47.5}} &  \footnotesize{\textbf{39.4}} \\

        \hspace{1.5mm} \footnotesize{Whispering-LlaMA~\cite{radhakrishnan2023whispering}} & \footnotesize{26M} & \footnotesize{90.8} &  \footnotesize{53.2} &  \footnotesize{101.1} &  \footnotesize{105.9} & \footnotesize{86.1} & \footnotesize{92.6} & \footnotesize{89.1} & \footnotesize{106.4} & \footnotesize{90.7} \\

        \hline

        \multicolumn{9}{l}{\footnotesize{\sysname-7B~~\textcolor{gray}{(w/ LLaMA2-7B)}}} \\
        
        \hspace{1.5mm} \footnotesize{w/ Whisper~(Large-v2)} & \footnotesize{17M} & \footnotesize{37.8} &  \footnotesize{18.2} &  \footnotesize{40.9} &  \footnotesize{37.5} & \footnotesize{18.6} & \footnotesize{37.0} & \footnotesize{40.1} & \footnotesize{44.9} & \footnotesize{34.4} \\

        \hspace{1.5mm} \footnotesize{w/ \textsc{LoRA} fine-tuned Whisper} & \footnotesize{17M} &  \cellcolor{green!20}\footnotesize{\textbf{34.6}~$\dagger$} & \footnotesize{17.6} &  \footnotesize{35.3} &  \cellcolor{green!20}\footnotesize{\textbf{35.6}~$\dagger$} & \footnotesize{18.2} & \footnotesize{34.8} &  \footnotesize{42.6} & \footnotesize{45.1} & \footnotesize{33.0} \\

        \multicolumn{9}{l}{\footnotesize{\sysname-13B~~\textcolor{gray}{(w/ LLaMA2-13B)}}} \\
        
        \hspace{1.5mm} \footnotesize{w/ Whisper~(Large-v2)} & \footnotesize{19M} & \footnotesize{37.1} &  \footnotesize{17.4} &  \footnotesize{40.4} &  \footnotesize{37.8} & \footnotesize{18.6} & \footnotesize{37.0} & \footnotesize{40.1} & \footnotesize{45.0} & \footnotesize{34.1} \\

        \hspace{1.5mm} \footnotesize{w/ \textsc{LoRa} fine-tuned Whisper} & \footnotesize{19M} &  \footnotesize{34.9} & \cellcolor{green!20} \footnotesize{\textbf{16.8}}~$\dagger$ & 
        \cellcolor{green!20} \footnotesize{\textbf{34.8}~$\dagger$} & \footnotesize{36.5} & \footnotesize{17.6} & \cellcolor{green!20} \footnotesize{\textbf{34.5}~$\dagger$} & \footnotesize{41.4} & \cellcolor{green!20} \footnotesize{\textbf{45.0}}~$\dagger$ & \cellcolor{green!20} \footnotesize{\textbf{32.7}~$\dagger$} \\

        \hline
        \hline
    \end{tabular}
}

\end{table*}

\begin{table*}[t!]
\centering
\caption{Comparison of the performance~(WER \%) of \textbf{multilingual} \sysname using LLaMA2-7B.}
\label{table:res_multi_asr}
\resizebox{\linewidth}{!}{
    \begin{tabular}{ l cccccccc|c }
        \hline
        \hline
        
        % \multicolumn{1}{c|}{\multirow{2}{*}{\textbf{\small{Method}}}} & \multicolumn{2}{c|}{\textbf{\footnotesize{Overall}}} & \multicolumn{2}{c|}{\textbf{\footnotesize{Hindi}}} & \multicolumn{2}{c|}{\textbf{\footnotesize{Gujarati}}} & \multicolumn{2}{c|}{\textbf{\footnotesize{Marathi}}} & \multicolumn{2}{c|}{\textbf{\footnotesize{Telugu}}} & \multicolumn{2}{c|}{\textbf{\footnotesize{Tamil}}} & \multicolumn{2}{c|}{\textbf{\footnotesize{Malayalam}}} & \multicolumn{2}{c}{\textbf{\footnotesize{Punjabi}}} \\

        \multicolumn{1}{c}{\textbf{\scriptsize{Method}}} & \textbf{\scriptsize{\# params}} & 
        \textbf{\scriptsize{Gujarati}} & 
        \textbf{\scriptsize{Hindi}} & 
        \textbf{\scriptsize{Malayalam}} & \textbf{\scriptsize{Marathi}} & 
        \textbf{\scriptsize{Punjabi}}  &
        \textbf{\scriptsize{Tamil}} &
        \textbf{\scriptsize{Telugu}} &   \textbf{\scriptsize{Average}} \\

        \hline

        \footnotesize{Whisper~(Large-v2)~\cite{whisper}} & --  &  \footnotesize{108.2} &  \footnotesize{35.9} &  \footnotesize{107.8} & \footnotesize{84.7} & \footnotesize{101.8} & \footnotesize{48.0} & \footnotesize{104.4} &
        \footnotesize{84.4}\\ 

        \hspace{2mm} \footnotesize{w/ \textsc{LoRA} fine-tuning~\cite{lora}} & \footnotesize{15M}  &  \footnotesize{56.8} &  \footnotesize{19.8} &  \cellcolor{green!20} \footnotesize{\textbf{39.8}~$\dagger$} & \footnotesize{37.4} &  \footnotesize{55.8} & \cellcolor{green!20} \footnotesize{\textbf{39.8}~$\dagger$} & \footnotesize{50.5}
        & \footnotesize{42.8}\\

        \footnotesize{\sysname-7B w/ Whisper~(Large-v2)} & \footnotesize{17M} & \cellcolor{green!20} \footnotesize{\textbf{36.4}~$\dagger$} & \cellcolor{green!20} \footnotesize{\textbf{17.5}~$\dagger$} & \footnotesize{40.0} & \cellcolor{green!20} \footnotesize{\textbf{36.7}~$\dagger$} & \cellcolor{green!20} \footnotesize{\textbf{36.1}~$\dagger$} & \footnotesize{40.2} &\cellcolor{green!20} \footnotesize{\textbf{43.2}~$\dagger$} &\cellcolor{green!20} \footnotesize{\textbf{35.8}~$\dagger$} \\

        \hline
    \end{tabular}
}

\end{table*}
\noindent \textbf{Models.} We use the Whisper Large-v2 encoder-decoder model for ASR (1.55B parameters). For the LLM, we experiment with two models of varying size, namely LLaMA2-7B and LLaMA2-13B. 
%These models are trained with 2 trillion tokens of publicly available data.
\sysname additionally uses $8$ projection layers that each first reduce the dimensionality down from $768$ (matching Whisper Large-v2's dimensionality) to $192$, use SiLU activations, and then project up from 192 to match the dimensionality of the corresponding LLaMA2 model~(i.e $4096$ for LLaMA2-7B and $5120$ for LLaMA2-13B). For all our \sysname experiments, we use Lit-GPT~\cite{lit_gpt} library to modify the LLM decoding to work in tandem with the ASR decoder. 
%\todo{Lit-GPT is unclear to me.}
%Finally, for the length model, we employ a simple linear regression architecture with three degrees of polynomial that is trained using the scikit-learn library.
%We run all our experiments using Lit-GPT~\cite{lit_gpt}, an open-source language modelling toolkit, on NVIDIA A-100 GPUs.

\begin{figure}[t!]
    \centering
    \includegraphics[width=0.48\textwidth]{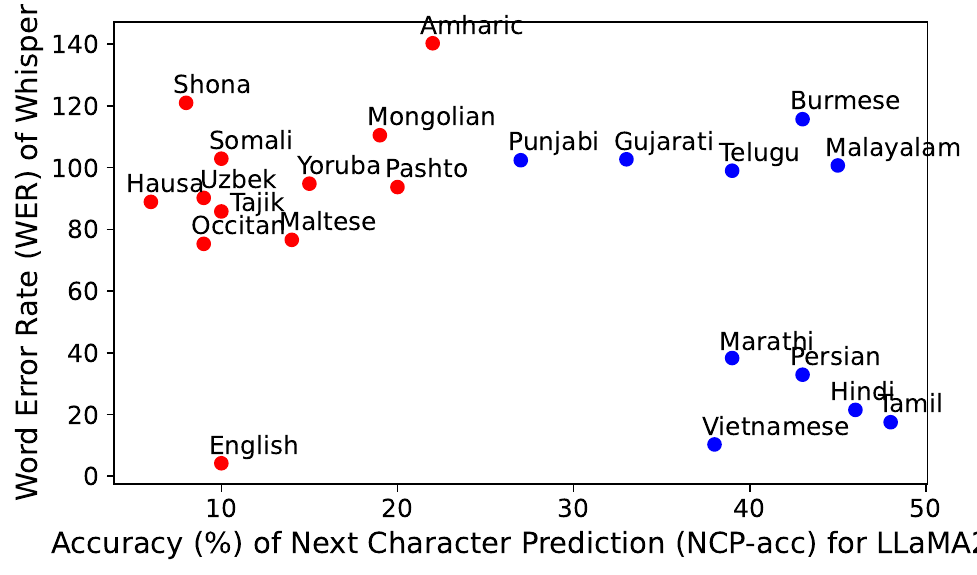}
    \caption{2D plot comparing the accuracy~(\%) of LLaMA2 on Next Character Prediction~(NCP-acc) with the Word Error Rate~(WER \%) of Whisper on a subset of languages from FLEURS. The plot serves as a point of reference for selecting languages that might benefit from \sysname. Specifically, we chose languages~(colored in \textcolor{blue}{blue}) that have high NCP-acc and medium to high WER using Whisper.}
    \label{fig:analysis:language_selection}
\end{figure}

\vspace{0.3em}
\noindent \textbf{Dataset.}  We evaluate on a subset of the FLEURS multilingual benchmark dataset~\cite{conneau2023fleurs}. This is an n-way parallel dataset that consists of roughly 12 hours of supervision for over 100 languages. Our conjecture is that \sysname could be beneficial in improving the ASR performance of those low-resource languages for which Whisper's ASR performance is poor, incurring moderate to high word error rates (WERs), and for which the LLM has reasonable text generation capabilities. The latter is quantitatively measured using the following metric. For a validation set of roughly 500 sentences in each evaluation language, we use the LLM to autoregressively predict only the next character given the past history. We loosely treat a single character across languages as having roughly similar durations in its underlying speech. For each evaluation language, we measure the fraction of next character predictions (NCP-acc) that exactly match the ground-truth characters. Figure~\ref{fig:analysis:language_selection} shows both Whisper's WERs and NCP-acc scores for a subset of FLEUR languages. We are interested in languages that appear in the upper and lower right quadrants. That is, those languages that incur moderate to high WERs and have fairly high NCP-acc scores. Based on this analysis, our final set of evaluation languages are Hindi, Gujarati, Marathi, Malayalam, Persian, Punjabi, Tamil and Telugu. We do not choose Burmese because severe under-tokenization results in most utterances exceeding Whisper decoder's fixed length of 448 tokens. We do not choose Vietnamese as the WER with pretrained Whisper is already quite good ($10.3\%$). 

\vspace{0.3em}
\noindent \textbf{Training and Inference.} For all our chosen evaluation languages, we train the model for $35$ epochs with a batch size of $32$, a learning rate of $0.001$ and a maximum of $2000$ steps. We use the AdamW~\cite{loshchilov2017decoupled} optimizer with a weight decay of $0.02$. To alleviate hallucinations during inference, we use nucleus sampling with top-p value of $0.9$ and top-k value of $10$. %We found this better than greedy decoding to address the issue of hallucination, which is a common occurrence in such large models. 
For some utterances on which the ASR performs very poorly, the LLM is prone to repetitions and gets stuck in a loop predicting up to $\text{max}_l$ tokens without predicting the end-of-sentence token. To mitigate this issue, we train a simple duration-to-length regressor for each language that takes the duration of the speech as its input and predicts an estimate of the number of output tokens. If the length of the LLM prediction far exceeds this predicted length, we truncate the overall prediction to the estimated length from the regressor. 

\vspace{0.3em}
\noindent \textbf{Baselines.} We compare \sysname with Whisper-v2 finetuned on the labeled data in each language using LoRA fine-tuning~\cite{lora}. We also compare against a recent  ASR-LLM fusion model Whispering-Llama~\cite{radhakrishnan2023whispering} that prompts the LLM with $n$-best lists from the Whisper ASR and trains the LLM via cross-attention adapter modules to attend to the Whisper encoder states.

\section{Experimental Results and Analysis}
Table~\ref{table:res_asr} shows the overall ASR results across eight languages comparing \sysname with the two baseline systems. Whispering-Llama underperforms due to the poor quality of the $n$-best lists derived from Whisper for these low-resource languages. \sysname can be used either with a pretrained or a finetuned Whisper ASR model. We are careful to train the same number of parameters using \sysname and LoRA finetuning. We observe that \sysname on top of a finetuned Whisper significantly outperforms the finetuned Whisper baseline by an overall relative WER reduction of $16\%$. This attests to \sysname's ability to derive complementary benefits over and above finetuning. The bigger Llama-13B model does not offer any consistent advantage over Llama-7B; the latter yields the best WERs for four test languages.  

%\subsection{Heuristics for choosing the evaluation languages}
%\todo{Here let us also talk about the case of Amheric.}

\vspace{0.5em}
\noindent \textbf{Multilingual \sysname.} 
We test how \sysname performs with multilingual data used to train projection layers shared across all languages. We pick seven of the Indian languages for this experiment. As shown in Table~\ref{table:res_multi_asr}, \sysname outperforms the multilingual fine-tuned Whisper model by $20\%$ signifying the superior adaptation capability of \sysname with similar number of parameters.
\looseness=-1
%While \sysname consistently improves upon the underlying Whisper model, we test the effect in a multilingual setting, where same set of parameters are used to adapt to seven different languages from the fleurs dataset. As shown in Table \ref{table:res_multi_asr}, we find that \sysname outperforms the multilingual finetuned Whisper model by 20\% signifying the superior adaptation capability with similar number of parameters.

\iffalse
\noindent \textbf{Effect of size of training data.} Figure~\ref{fig:anlaysis:dataset_size} shows how WERs vary as a function of the amount of Gujarati finetuning data available for both Whisper finetuning and \sysname-7B. \sysname appears to benefit more from a small fraction of the labeled data (going from 4 to 5 hours) compared to Whisper finetuning, after which both techniques exhibit the same WER trend with \sysname maintaining a constant (and significant) lead.
\begin{figure}[h!]
    \centering
    \includegraphics[width=0.4\textwidth]{figures/frac_analysis.pdf}
    \caption{Comparison of the performance of Whisper LoRa finetuning with \sysname on varying amounts of Gujarati dataset from FLEURS. x-axis represents the fraction of training dataset used and y-axis represents WER on test set.}
    \label{fig:anlaysis:dataset_size}
\end{figure}
\fi

\vspace{0.5em}
\noindent \textbf{Ablation Analysis.} 
We study the effect of the number of adaptation parameters and the positions of adaptation layers on \sysname, as shown in Table \ref{table:res_ablation}. We observe that using 8 adapter layers all at the end of the decoder yields almost similar performance as the model which has adaptation layers uniformly distributed. Reducing the number of adaptation layers for both \sysname-7B and \sysname-13B results in a degradation of around relative 5\% WER, but is still significantly better than LoRA finetuning which yields an average WER of $39.4\%$. This shows that \sysname can adapt and generalize much better than existing baselines even with a small number of parameters.

\begin{table}[t!]
\centering
\caption{Ablation study comparing the number of adapter layers and the position of adapter layers averaged across 8 languages.}
\label{table:res_ablation}
\resizebox{\columnwidth}{!}{
    \begin{tabular}{ l rc }
        \hline
        \hline
        
        % \multicolumn{1}{c|}{\multirow{2}{*}{\textbf{\small{Method}}}} & \multicolumn{2}{c|}{\textbf{\footnotesize{Overall}}} & \multicolumn{2}{c|}{\textbf{\footnotesize{Hindi}}} & \multicolumn{2}{c|}{\textbf{\footnotesize{Gujarati}}} & \multicolumn{2}{c|}{\textbf{\footnotesize{Marathi}}} & \multicolumn{2}{c|}{\textbf{\footnotesize{Telugu}}} & \multicolumn{2}{c|}{\textbf{\footnotesize{Tamil}}} & \multicolumn{2}{c|}{\textbf{\footnotesize{Malayalam}}} & \multicolumn{2}{c}{\textbf{\footnotesize{Punjabi}}} \\

        \multicolumn{1}{c}{\textbf{\scriptsize{Method}}} & \textbf{\scriptsize{\# params}} & \textbf{\scriptsize{Average WER}} \\

        \hline

        % \footnotesize{Whisper~(Large-v2)~\cite{whisper}} & -- & \footnotesize{64.4} &  \footnotesize{21.5} &  \footnotesize{102.7} &  \footnotesize{38.3} & \footnotesize{99.0} & \footnotesize{17.5} & \footnotesize{100.7} & \footnotesize{102.4} & \footnotesize{32.9} \\

        % \hline

        \multicolumn{3}{l}{\footnotesize{\sysname-7B~~\textcolor{gray}{(w/ LLaMA2-7B)}}} \\

        \hspace{1.5mm} \footnotesize{8 adapter layers (uniformly distributed)} & \footnotesize{17M} & \footnotesize{34.4} \\

         \hspace{1.5mm} \footnotesize{8 adapter layers (all at the end)} & \footnotesize{17M} & \footnotesize{34.8} \\

        \hspace{1.5mm} \footnotesize{4 adapter layers (uniformly distributed)} & \footnotesize{8.5M} & \footnotesize{36.4} \\

        \hspace{1.5mm} \footnotesize{4 adapter layers (all at the end)} & \footnotesize{8.5M} & \footnotesize{36.9} \\

        \hline
    \end{tabular}
}

\end{table}

\vspace{0.5em}
\noindent \textbf{Runtime Complexity.} We compare the real-time factor (RTF) of \sysname with the original Whisper model and Whispering-Llama model. That is, the amount of time it takes to decode 1 sec of audio. RTFs of Whisper, \sysname and Whispering-Llama on an A100 80GB GPU are 0.42 secs, 0.64 secs and 1.3 secs, respectively. For the Whispering-Llama model, a significant fraction of the total time ($>60\%$) is spent in generating $n$-best lists which adversely affects the overall RTF. For training, \sysname takes around 1 hour on an A100 80GB GPU, while Whispering-Llama takes around 6 hours owing to high training overhead in generating $n$-best lists and learning expensive cross-attention over significantly long audio embeddings. LoRA finetuning takes roughly the same time as Whispering-Llama on a smaller A100 40GB GPU (as there are no Llama2 weights to load). SALSA is much faster to train in comparison to the baselines owing to its simple architecture.
%

%versus the proportion of the dataset used for training, for both Whisper Fine-tuned model and our proposed \sysname-7B model. We conduct this analysis on the Gujarati language of the \textsc{Fleurs} dataset, aiming to assess the effectiveness of these models in low-resource to extremely low-resource scenario. First, we observe that, across all the training data scenarios, our model is consistently better than Whisper. Furthermore, we observe that the rate of improvement, as indicated by the slope of the plots, is significantly higher for \sysname compared to Whisper. This suggests that even with a small amount of additional data, \sysname is capable of adapting more rapidly as compared to other architectures.

\section{Conclusion}
In this paper we presented \sysname, a light-weight fusion of an ASR system with an LLM that retains the ASR model's expertise in encoding and decoding audio, while harnessing the superior language modeling capabilities of the LLM.  Our method provides significant reductions in WER compared to fine-tuning the ASR model alone, while providing efficient one-pass decoding, and much faster training than existing LLM-ASR fusion methods.  In this paper, we focused on improving the transcription of isolated utterances. In future, we plan to harness the instruction-following capabilities of LLMs for more applications that require stateful contextual biasing.

% \secate grtion{Outline}
% \begin{itemize}
%     \item Limitations: Latency of decoding, and effort required for finetuning (time and compute). Existing approaches: Tight coupling (attention over encoder), nbest integration (Whispering llama) and speech-GPT like approaches.  
%     \item Our approach: Simple projection layers on the last decoder state -- couple pretrained LLM and ASR. 
%     \item Direct decoding from the LLM; no two-pass overhead. 
%     \item Leveraging the decoder of the ASR; there's a vocab mismatch -- which we handle with a cascaded tokenization. 
%     \item Fine-tune the ASR model. All of that complements our approach. 
% \end{itemize}

\bibliographystyle{IEEEtran}
\bibliography{mybib}

\end{document}